\documentclass[11pt,a4paper]{article}
\usepackage{times,latexsym}
\usepackage{url}
\usepackage[T1]{fontenc}
\usepackage{adjustbox}
\usepackage{algorithm}
\usepackage{algorithmicx}
\usepackage{amssymb}
\usepackage{amsmath}
\usepackage{amsthm}
\usepackage{booktabs}
\usepackage{graphicx}
\usepackage{makecell}
\usepackage{multirow}
\usepackage{multicol}
\usepackage{subfig}
\usepackage{tabularx}
\usepackage{caption}
\usepackage[normalem]{ulem}
\useunder{\uline}{\ul}{}
\usepackage[shortlabels]{enumitem}
\usepackage[utf8]{inputenc}

\usepackage[acceptedWithA]{tacl2018v2}
\usepackage{mathbbol}

\newtheorem{definition}{Definition}
\newcommand{\com}[1]{}

\usepackage{xspace,mfirstuc,tabulary}

\newif\iftaclinstructions
\taclinstructionsfalse 
\iftaclinstructions
\renewcommand{\confidential}{}
\renewcommand{\anonsubtext}{(No author info supplied here, for consistency with
TACL-submission anonymization requirements)}
\newcommand{\instr}
\fi

\iftaclpubformat 

\else

\fi

\title{Model Compression for Domain Adaptation through Causal Effect Estimation}

\author{Guy Rotman\thanks{ \hspace{1mm} Authors contributed equally.} , Amir Feder\footnotemark[1] , Roi Reichart\\
Faculty of Industrial Engineering and Management, Technion, IIT\\
\texttt{grotman@campus.technion.ac.il} \\
\texttt{feder@campus.technion.ac.il} \\
\hspace{1.2em}\texttt{roiri@technion.ac.il} \\
}

\begin{document}
\maketitle
\begin{abstract}
Recent improvements in the predictive quality of natural language processing systems are often dependent on a substantial increase in the number of model parameters. This has led to various attempts of compressing such models, but existing methods have not considered the differences in the predictive power of various model components or in the generalizability of the compressed models. To understand the connection between model compression and out-of-distribution generalization, we define the task of compressing language representation models such that they perform best in a domain adaptation setting. We choose to address this problem from a causal perspective, attempting to estimate the \textit{average treatment effect} (ATE) of a model component, such as a single layer, on the model's predictions. Our proposed ATE-guided Model Compression scheme (\textbf{AMoC}), generates many model candidates, differing by the model components that were removed. Then, we select the best candidate through a stepwise regression model that utilizes the ATE to predict the expected performance on the target domain. \textbf{AMoC} outperforms strong baselines on dozens of domain pairs across three text classification and sequence tagging tasks. \footnote{Our code and data are available at: \url{https://github.com/rotmanguy/AMoC}.} \footnote{This is a pre-MIT Press publication version.}
\end{abstract}

\section{Introduction}
\label{sec:intro}

The rise of deep neural networks (DNNs) has transformed the way we represent language, allowing models to learn useful features directly from raw inputs. However, recent improvements in the predictive quality of language representations are often related to a substantial increase in the number of model parameters. Indeed, the introduction of the Transformer architecture~\cite{vaswani2017attention} and attention-based models ~\cite{devlin2019bert, liu2019roberta, brown2020language} has improved performance on most natural language processing (NLP) tasks, while facilitating a large increase in model sizes.


Since large models require a significant amount of computation and memory during training and inference, there is a growing demand for compressing such models while retaining the most relevant information. While recent attempts have shown promising results \cite{sanh2019distilbert}, they have some limitations. Specifically, they attempt to mimic the behavior of the larger models without trying to understand the information preserved or lost in the compression process.

In compressing the information represented in billions of parameters, we identify three main challenges. First, current methods for model compression are not interpretable. While the importance of different model parameters is certainly not uniform, it is hard to know \textit{a-priori} which of the model components should be discarded in the compression process. This notion of feature importance has not yet trickled down into compression methods, and they often attempt to solve a dimensionality reduction problem where a smaller model aims to mimic the predictions of the larger model. Nonetheless, not all parameters are born equal, and only a subset of the information captured in the network is actually useful for generalization \cite{frankle2018lottery}. 

The second challenge we observe in model compression is out-of-distribution generalization. Typically, compressed models are tested for their in-domain generalization. However, in reality the distribution of examples often varies and is different than that seen during training. Without testing for the generalization of the compressed models on different test-set distributions, it is hard to fully assess what was lost in the compression process. The setting explored in domain adaptation provides us with a platform to test the ability of the compressed models to generalize across-domains, where some information that the model has learned to rely on might not exist. Strong model performance across domains provides a stronger signal on retaining valuable information. 

Lastly, another challenge we identify in training and selecting compressed models is confidence estimation. In trying to understand what gives large models the advantage over their smaller competitors, recent probing efforts have discovered that commonly used models such as BERT \cite{devlin2019bert}, learn to capture semantic and syntactic information in different layers and neurons across the network \cite{rogers2020primer}. While some features might be crucial for the model, others could learn spurious correlations that are only present in the training set and are absent in the test set~\cite{kaushik2019learning}. Such cases have led to some intuitive common practices such as keeping only layers with the same parity or the top or bottom layers \cite{fan2019reducing, sajjad2020poor}. Those practices can be good on average, but do not provide model confidence scores or success rate estimates on unseen data. 

Our approach addresses each of the three main challenges we identify, as it allows estimating the marginal effect of each model component, is designed and tested for out-of-distribution generalization, and provides estimates for each compressed model performance on an unlabeled target domain. We dive here into the connection between model compression and out-of-distribution generalization, and ask whether compression schemes should consider the effect of individual model components on the resulting compressed model. Particularly, we present a method that attempts to compress a model while maintaining components that can generalize well across domains. 

Inspired by causal inference \cite{pearl1995causal}, our compression scheme is based on estimating the average effect of model components on the decisions the model makes, at both the source and target domains. In causal inference, we measure the effect of interventions by comparing the difference in outcome between the control and treatment groups. In our setting, we take advantage of the fact that we have access to unlabeled target examples, and treat the model's predictions as our outcome variable. We then try to estimate the effect of a subset of the model components, such as one or more layers, on the model's output. 


To do that, we propose an approximation of a counterfactual model where a model component of choice is removed. We train an instance of the model without that component and keep everything else equal apart from the input and output to that component, which allows us to perform only a small number of gradient steps. Using this approximation, we then estimate the \textit{average treatment effect} (ATE) by comparing the predictions of the base model to those of its counterfactual instance.

Since our compressed models are very efficiently trained, we can generate a large number of such models per each source-target domain pair. We then train a regression model on our training domain pairs in order to predict how well a compressed model would generalize from a source to a target domain, using the ATE as well as other variables. This regression model can then be applied to new source-target domain pairs in order to select the compressed model that best supports cross-domain generalization.

To organize our contributions, we formulate three research questions:
\vspace{-5pt}
\begin{enumerate}
\setlength{\itemsep}{-4pt}
    \item Can we produce a compressed model that outperforms all baselines in out-of-distribution generalization?
    \item Does the model component we decide to remove indeed hurt performance the least? 
    \item Can we use the average treatment effect to guide our model selection process?
\end{enumerate}

In $\S$ \ref{sec:results} we directly address each of the three research questions, and demonstrate the usefulness of our method, ATE-guided model compression (\textbf{AMoC}), to improve model generalization. 

\section{Previous Work}
Previous work on the intersection of neural model compression, domain adaptation and causal inference is limited, as our application of causal inference to model compression and our discussion of the connection between compression and cross-domain generalization are novel. However, there is an abundance of work in each field on its own, and on the connection between domain adaptation and causal inference. Since our goal is to explore the connection between compression and out-of-distribution generalization, as framed in the setting of domain adaptation, we survey the literature on model compression and the connection between generalization, causality and domain adaptation.

\subsection{Model Compression}

NLP models have been increased exponentially in size, growing from less than a million parameters a few years ago to hundreds of billions. Since the introduction of the Transformer architecture, this trend has been strengthened, with some models reaching more than 175 billion parameters \cite{brown2020language}. As a result, there has been a growing interest in compressing the information captured in Transformers into smaller models \cite{chen2020adabert, ganesh2020compressing, sun2020mobilebert}. 

Usually, such smaller models are trained using the base model as a teacher, with the smaller student model learning to predict its output probabilities \cite{hinton2015distilling, jiao2019tinybert, sanh2019distilbert}. However, even if the student closely matches the teacher's soft labels, their internal representations may be considerably different. This internal mismatch can undermine the generalization capabilities originally intended to be transferred from the teacher to the student \cite{aguilar2020knowledge, mirzadeh2020improved}. 

As an alternative, we try not to interfere or alter the learned representation of the model. Compression schemes such as those presented in \citet{sanh2019distilbert} discard model components randomly. Instead, we choose to focus on understanding which components of the model capture the information that is most useful for it to perform well across domains, and hence should not be discarded.


\subsection{Domain Adaptation and Causality}
Domain adaptation is a longstanding challenge in machine learning (ML) and NLP, which deals with cases where the train and test sets are drawn from different distributions. A great effort has been dedicated to exploit labels from both source and target domains for that purpose \cite{daume2010frustratingly, sato2017adversarial, cuy2018transfer, lin2018neural, wang2018label}. However, a much more challenging and realistic scenario, also termed as \textit{unsupervised domain adaptation}, occurs when no labeled target samples exist \cite{blitzer2006domain, ganin2016domain, ziser2017neural, ziser2018deep, ziser2018pivot, ziser2019task, rotman2019deep, ben2020perl}. In this setting, we have access to labeled and unlabeled data from the source domain and to unlabeled data from the target, and models are tested by their performance on unseen examples from the target domain.

A closely related task is domain adaptation success prediction. This task explores the possibility of predicting the expected performance degradation between source and target domains \cite{mcclosky2010automatic, elsahar2019annotate}. Similar to predicting performance in a given NLP task, methods for predicting domain adaptation success often rely on in-domain performance and distance metrics estimating the difference between the source and target distributions \cite{reichart2007ensemble, ravi2008automatic, nenkova2009performance, van2010using, xia2020predicting}. While these efforts have demonstrated the importance of out-of-domain performance prediction, they have not been made as far as we know in relation to model compression.

As the fundamental purpose of domain adaptation algorithms is improving the out-of-distribution generalization of learning models, it is often linked with causal inference \cite{johansson2016learning}. In causal inference we typically care about estimating the effect that an intervention on a variable of interest would have on an outcome \cite{pearl2009causality}. Recently, using causal methods to improve the out-of-distribution performance of trained classifiers is gaining traction \cite{rojas2018invariant, wald2021calibration}. 

Indeed, recent papers applied a causal approach to domain adaptation. Some researchers proposed using causal graphs to predict under distribution shifts \cite{scholkopf2012causal} and to understand the type of shift \cite{zhang2013domain}. Adapting these ideas to computer vision, \citet{gong2016domain} were one of the first to propose a causal graph describing the generative process of an image as being generated by a “domain”. The causal graph served for learning invariant components that transfer across domains. Since that, the notion of invariant prediction has emerged as an important operational concept in causal inference \cite{peters2017elements}. This idea has been used to learn classifiers that are robust to domain shifts and can perform well on unseen target distributions \cite{gong2016domain, magliacane2018domain, rojas2018invariant, greenfeld2019robust}. 

Here we borrow ideas from causality to help us reason on the importance of specific model components, such as individual layers. That is, we estimate the effect of a given model component (denoted as the \textit{treatment}) on the model's predictions in the unlabeled target domain, and use the estimated effect as an evaluation of the importance of this component. Our treatment effect estimation method is inspired by previous causal model explanation work \cite{goyal2019explaining, feder2021causalm}, although our algorithm is very different.

\section{Causal Terminology}
Causal methodology is most commonly used in cases where the goal is estimating effects on real-world outcomes, but it can be adapted to help us understand and explain what affects NLP models \cite{feder2021causalm}. Specifically, we can think of intervening on a model and altering its components as a causal question, and measure the effect of this intervention on model predictions. A core benefit of this approach is that we can estimate treatment effects on model's predictions without the need for manually-labeled target data.

Borrowing causal methodology into our setting, we treat model components as our treatment, and try to estimate the effect of removing them on our model's predictions. The predictions of a model are driven by its components, and by changing one component and holding everything else equal, we can estimate the effect of this intervention. We can use this estimation in deciding which model component should be kept in the compression process. 

As the link between model compression and causal inference was not explored previously, we provide here a short introduction to causal inference and its basic terminology, focusing on its application to our use case. We then discuss the connection to Pearl's \emph{do}-operator~\cite{pearl2009causal} and the estimation of treatment effects.

Imagine we have a model $m$ that classifies examples to one of $L$ classes. Given a set $\mathcal{C}$ of $K$ model components, which we hypothesize might affect the model's decision, we denote the set of binary variables $I_c = \{I_{c_j} \in \{0,1\} | j \in \{1, \ldots, K \} \}$, where each corresponds to the inclusion of the component in the model, i.e., if $I_{c_j}=1$ then the $j$-th component ($c_j$) is in the model. Our goal is to assert how the model's predictions are affected by the components in $\mathcal{C}$. As we are interested in the effect on the class probability assigned by $m$, we measure this probability for an example $x$, and denote it for a class $l$ as $z(m(x))_{l}$ and for all $L$ classes as $\vec{z}(m(x))$.

Using this setup, we can now define the ATE, the common metric used when estimating causal effects. ATE is the difference in mean outcomes between the treatment and control groups, and using \emph{do}-calculus~\cite{pearl1995causal} we can define it as follows:
\begin{definition}[Average Treatment Effect (ATE)] 
The average treatment effect of a binary treatment $I_{c_j}$ on the outcome $\vec{z}(m(x))$ is:
\begin{equation}\label{def:ATE}
\begin{aligned}
    \text{ATE} ({c_j}) = & \mathbb{E}\left[\vec{z}(m(x))|do(I_{c_j}=1)\right] \\
    & - \mathbb{E}\left[\vec{z}(m(x))|do(I_{c_j}=0)\right],
\end{aligned}
\end{equation}
\end{definition}
where the \textit{do}-operator is a mathematical operator introduced by \citet{pearl1995causal}, which indicates that we intervene on $c_j$ such that it is included ($do(I_{c_j}=1)$) or not ($do(I_{c_j}=0)$) in the model. 

While the setup usually explored with \emph{do}-calculus involves a fixed joint-distribution where treatments are assigned to individuals (or examples), we borrow intuition from a specialized case where interventions are made on the process which  generates outcomes given examples. This type of an intervention is called \textit{Process Control}, and was proposed by \citet{pearl2009causal} and further explored by \citet{bottou2013counterfactual}. This unique setup is designed to improve our understanding of the behavior of complex learning systems and predict the consequences of changes made to the system. Recently, \citet{feder2021causalm} used it to intervene on language representation models, generating a counterfactual representation model through an adversarial training algorithm which biases the representation model to forget information about treatment concepts and maintain information about control concepts.

In our approach we intervene on the $j$-th component, by holding the rest of the model fixed and training only the parameters that control the input and output to that component. This is crucial for our estimation procedure as we want to know the effect of the $j$-th component on a specific model instance. This effect can be computed by comparing the predictions of the original model instance to those of the intervened model (see below). This computation is fundamentally different from measuring the conditional probability where the $j$-th component is not in the model by estimating $\mathbb{E}\left[\vec{z}(m(x))|I_{c_j}=0\right]$.

\section{Methodology}
We start by describing the task of compressing models such that they perform well on out-of-distribution examples, detailing the domain adaptation framework we focus on. Then, we describe our compression scheme, designed to allow us to approximate the ATE and responsible for producing compressed model candidates. Finally, we propose a regression model that uses the ATE and other features to predict a candidate model's performance on a target domain. This regression allows us to select a strong candidate model.

\subsection{Task Definition and Framework}
To test the ability of a compressed model to generalize on out-of-distribution examples, we choose to focus on a domain adaptation setting. An appealing property of domain adaptation setups is that they allow us to measure out-of-distribution performance in a very natural way by training on one domain and testing on another. 

In our setup, during training, we have access to $n$ source-target domain pairs $(\mathbf{S}^i, \mathbf{T}^i)_{i=1}^{n}$. For each pair we assume to have labeled data from the source domains $(\mathbf{L_{S^{i}}})_{i=1}^{n}$ and unlabeled data from the the source and target domains $(\mathbf{U_{S^{i}}}$, $\mathbf{U_{T^{i}}})_{i=1}^{n}$. We also assume to have held-out labeled data for all domains, for measuring test performance $(\mathbf{H_{S^{i}}}, \mathbf{H_{T^{i}}})_{i=1}^{n}$. At test time we are given an unseen domain pair $(\mathbf{S^{n+1}}, \mathbf{T^{n+1}})$ with labeled source data $\mathbf{L_{S^{n+1}}}$ and unlabeled data from both domains $\mathbf{U_{S^{n+1}}}$ and $\mathbf{U_{T^{n+1}}}$, respectively. Our goal is to classify examples on the unseen target domain $\mathbf{T^{n+1}}$ using a compressed model $m^{n+1}$ trained on the new source domain.

For each domain pair in $(\mathbf{S^i}, \mathbf{T^i})_{i=1}^{n}$, we generate a set of $K$ candidate models $M^i = \{m_1^i,\ldots ,m_K^i\}$, differing by the model components that were removed from the base model $m^{i}_{B}$. For each candidate, we compute the ATE and other relevant features which we discuss in $\S$ \ref{subsec:features}. Then, using the training domain pairs, for which we have access to a limited amount of labeled target data, we train a stepwise linear regression to predict the performance of all candidate models in $\{M^i\}_{i=1}^{n}$ on their target domain. Finally, at test time, after computing the regression features on the unseen source-target pair, we use the trained regression model to select the compressed model $(m^{n+1})^{*} \in M^{n+1}$ that is expected to perform best on the unseen unlabeled target domain.

While this task definition relies on a limited number of labeled examples from some target domains at training time, at test time we only use labeled examples from the source domain and unlabeled examples from the target. We elaborate on our compression scheme, responsible for generating the compressed model candidates in $\S$ \ref{subsec:causal_compression}. We then describe the regression features and the regression model in $\S$ \ref{subsec:features} and $\S$ \ref{subsec:reg}, respectively.

\begin{figure*}[!ht]
    \centering
    \subfloat[\centering Base Model]
    {{\includegraphics[width=0.44\linewidth]{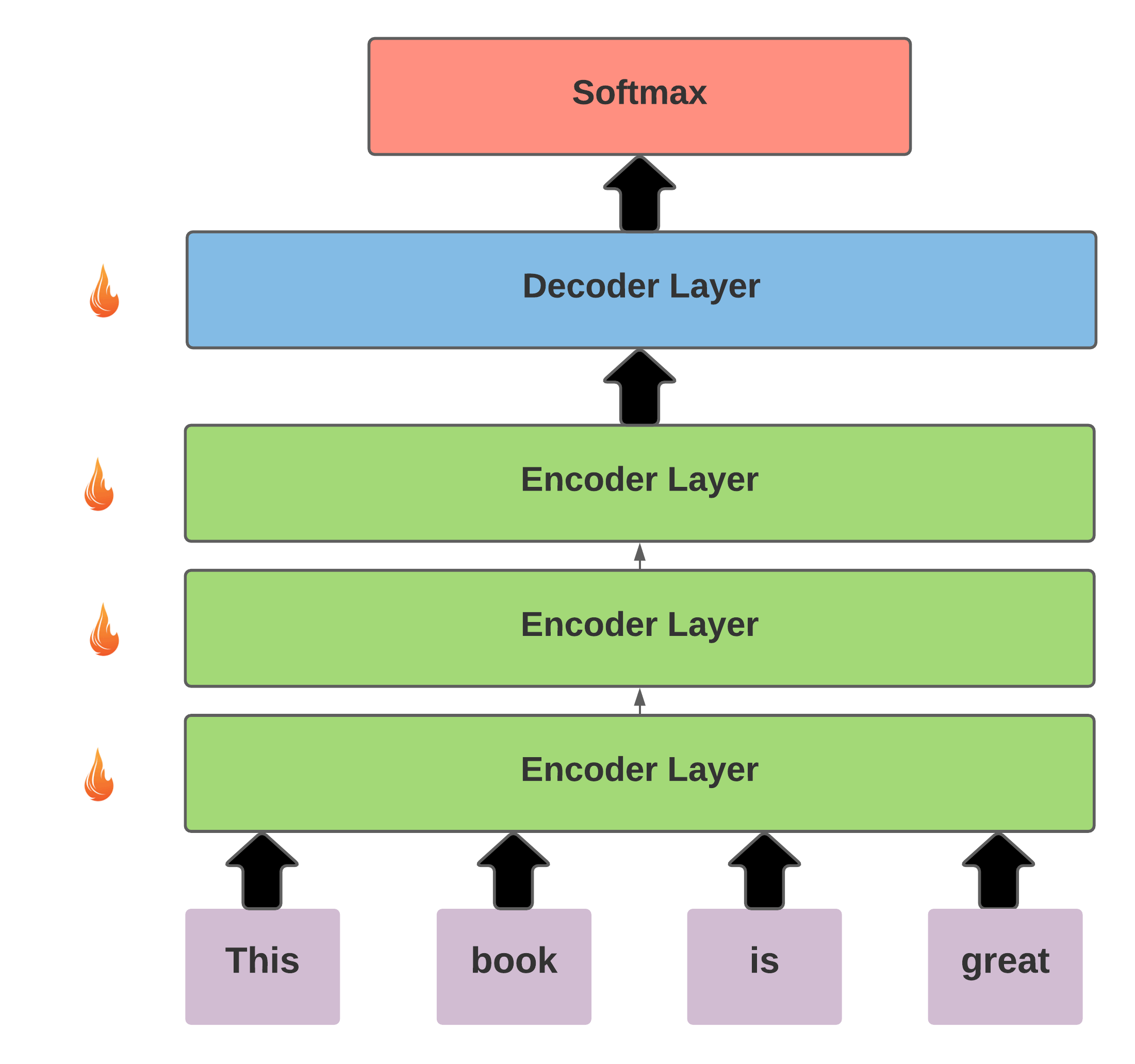}}}
    \qquad
    \centering
    \subfloat[\centering Compressed Model]
    {{\includegraphics[width=0.45\linewidth]{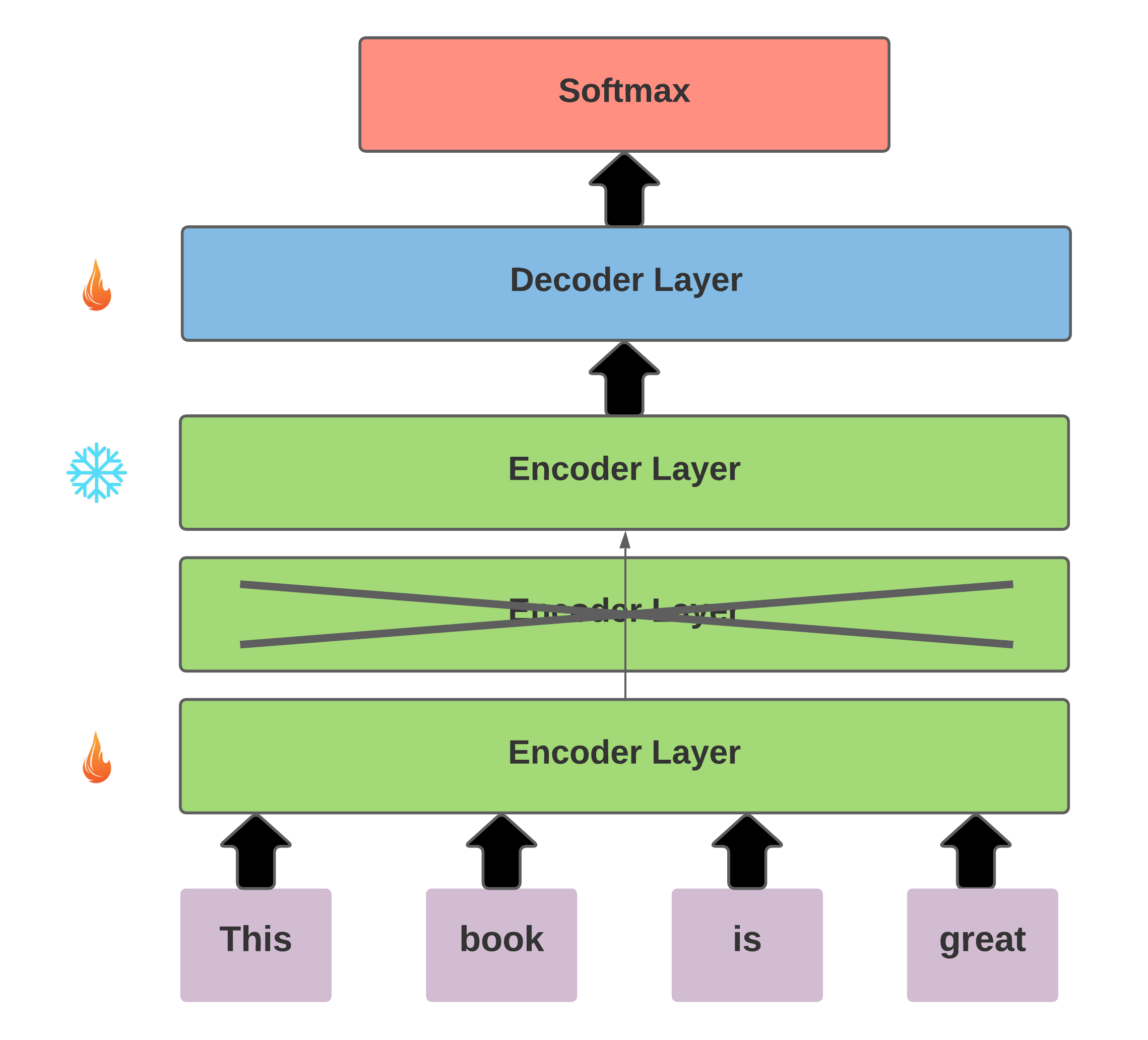}}}
    \caption{An example of our method with a 3-layer encoder when considering the removal of layer components. (a) At first, the base model is trained (Alg. \ref{alg:causal_compression}, step $1(a)$). (b) The second encoder layer is removed from the base model, and the first layer is connected to the final encoder layer. The compressed model is then fine-tuned for one or more epochs, where only the parameters of the first layer and the decoder are updated (Alg. \ref{alg:causal_compression}, step $1(b)$). We mark frozen layers and non-frozen layers with snow-flakes and fire symbols, respectively.} 
    \label{fig:model_architecture}
\end{figure*}

\subsection{Compression Scheme}
\label{subsec:causal_compression}
Our compression scheme (\textbf{AMoC}) assumes to operate on a large classifier, consisting of an encoder-decoder architecture, that serves as the base model being compressed. In such models, the encoder is the language representation model (e.g., BERT), and the decoder is the task classifier. Each input sentence $x$ to the base model $m^{i}_{B}$ is encoded by the encoder $e$. Then, the encoded sentence $e(x)$ is passed through the decoder $d$ to compute a distribution over the the label space $L$: $\vec{z}(m^{i}_{B}(x)) = Softmax(d(e(x)))$. \textbf{AMoC} is designed to remove a set of encoder components, and can in principle be used with any language encoder.

As described in Algorithm \ref{alg:causal_compression}, \textbf{AMoC} generates candidate compressed versions of $m^{i}_{B}$. In each iteration it selects from $\mathcal{C}$, the set containing subsets of encoder components, a candidate $c_{k} \in \mathcal{C}$ to be removed. \footnote{For example, if components correspond to layers, and we wish to remove an individual layer from a 12-layer encoder, then $\mathcal{C} = \{\{i\} | i\in \{1,\ldots, 12\}\}$.} The goal of this process is to generate many compressed model candidates, such that the $k$-th candidate $c_{k}$ differs from the base model $m^{i}_{B}$ only by the effect of the parameters in $c_{k}$ on the model's predictions. After generating these candidates, \textbf{AMoC} tries to choose the best performing model for the unseen target domain.

\begin{algorithm}[!t]
\caption{\footnotesize ATE-Guided Model Compression (AMoC)} \label{alg:causal_compression}
{\footnotesize
\textbf{Input:} Domain pairs $(\mathbf{S^i}, \mathbf{T^i})_{i=1}^{n+1}$ with Labeled source data $\mathbf{(L_{S^{i}})_{i=1}^{n+1}}$, Unlabeled source and target data $(\mathbf{U_{S^i}}, \mathbf{U_{T^i}})_{i=1}^{n+1}$, Labeled held-out source and target data $(\mathbf{H_{S^i}}, \mathbf{H_{T^i}})_{i=1}^{n}$, and a set $\mathcal{C}$ of subsets of encoder components to be removed.\\
$\textbf{Algorithm:}$
\begin{enumerate}
    \item For each domain pair in $(\mathbf{S^i}, \mathbf{T^i})_{i=1}^{n}$
    \begin{enumerate}
        \item Train the base model $m^{i}_{B}$ on $\mathbf{L_{S^i}}$.
            \item For $c_{k}\in \mathcal{C}$
                \subitem - Freeze all encoder parameters.
                \subitem - Remove every component in $c_{k}$ from $m^{i}_{B}$.
                \subitem - Connect and unfreeze the remaining components according to $\S$ \ref{subsec:causal_compression}.
                \subitem - Fine-tune the new model $m^{i}_{k}$ on $\mathbf{L_{S^i}}$ for one or more epochs.
                \subitem - Compute $\widehat{ATE}_{S^{i}}(c_{k})$ and $\widehat{ATE}_{T^i}(c_{k})$ according to Eq. \ref{eq:ate}, using $\mathbf{U_{S^i}}$ and $\mathbf{U_{T^i}}$. 
                \subitem - Compute the remaining features in \ref{subsec:features}.
    \end{enumerate}
    \item Train the stepwise regression according to Eq. \ref{eq:reg}, using all compressed models generated in step 1.
    \item Repeat steps 1(a)-1(b) for $(\mathbf{S^{n+1}}, \mathbf{T^{n+1}})$ and choose $(m^{n+1})^{*}$ with the highest expected performance according to the regression model.
\end{enumerate}
}
\end{algorithm}

When generating the $k$-th compressed model of the $i$-th source-target pair, we start by removing all parameters in $c_{k}$ from the computational graph of $m^{i}_{B}$. Then, we connect the predecessor of each detached component from $c_{k}$ to its successor in the graph, which yields the new ${m^{i}_{k}}$ (see Figure \ref{fig:model_architecture}). To estimate the effect of $c_{k}$ on the predictions of $m^{i}_{B}$, we freeze all remaining model parameters in ${m^{i}_{k}}$ and fine-tune it for one or more epochs, training only the decoder and the parameters of the new connections between the predecessors and successors of the removed components. An advantage of this procedure is that we can efficiently generate many model candidates. Figure \ref{fig:model_architecture} demonstrates this process on a simple architecture when considering the removal of layer components. 

Guiding our model selection step is the ATE of $c_{k}$ on the base model $m^{i}_{B}$. The generation of each compressed candidate ${m^{i}_{k}}$ is designed to allow us to estimate the effect of $c_{k}$ on the model's predictions. In comparing the predictions of $m^{i}_{B}$ to the compressed model ${m^{i}_{k}}$ on many examples, we try to mimic the process of generating control and treatment groups. As is done in controlled experiments, we compare examples that are given a treatment, i.e., encoded by the compressed model ${m^{i}_{k}}$, and examples that were encoded by the base model $m^{i}_{B}$. Intervening on the example-generating process was explored previously in the causality literature by \citet{bottou2013counterfactual,feder2021causalm}.

Alongside the ATE, we compute other features that might be predictive of a compressed model's performance on an unlabeled target domain, which we discuss in detail in $\S$ \ref{subsec:features}. Using those features and the ATE, we train a linear stepwise regression  to predict a compressed model's performance on target domains ($\S$ \ref{subsec:reg}). Finally, at test time \textbf{AMoC} is given an unseen domain pair and  applies the regression in order to choose the compressed source model expected to perform best on the target domain. Using the regression, we can estimate the power of the ATE in predicting model performance and answer Question $3$ of $\S$ \ref{sec:intro}.

In this paper, we choose to focus on the removal of sets of layers, as done in previous work \cite{fan2019reducing, sanh2019distilbert, sajjad2020poor}. While our method can support any other parameter partitioning, such as clusters of neurons, we leave this for future work. In the case of layers, to establish the new compressed model we simply connect the remained layers according to their hierarchy. For example, for a base model with a 12-layer encoder and $c=\{2,3,7\}$ the unconnected components are $\{1\}, \{4,5,6\}$ and $\{8,9,10,11,12\}$. Layer $1$ will then be connected to layer $4$, and layer $6$ to layer $8$. The compressed model will be then trained for one or more epochs where only the decoder and layers $1$ and $6$ (using the original indices) are fine-tuned . In times where layer 1 is removed, the embedding layer is connected to the first unremoved layer and is fine-tuned.

\subsection{Regression Features}
\label{subsec:features}

Apart from the ATE, which estimates the impact of the intervention on the base model, we naturally need to consider other features. Indeed, without any information on the target domain, predicting that a model will perform the same as in the source domain could be a reasonable first-order approximation \cite{mcclosky2010automatic}. Also, adding information on the distance between the source and target distributions \cite{van2010using} or on the type of components that were removed (such as the number of layers) might also be useful for predicting the model's success. We present here all the features we consider, and discuss their usefulness in predicting model performance. To answer Q$3$, we need to show that given all this information, the ATE is still predictive for the model's performance in the target domain.

\paragraph{ATE}
Our main variable of interest is the \textit{average treatment effect} of the components in $c_{k}$ on the predictions of the model. In our compression scheme, we estimate for a specific domain $d \in \{S^i, T^i\}$ the ATE for each compressed model ${m^{i}_{k}}$ by comparing it to the base model $m^{i}_{B}$:

\begin{equation} \label{eq:ate}
\widehat{ATE}_d(c_{k}) = \frac{1}{|\mathbf{U_d}|} \sum_{x \in \mathbf{U_d}} \langle \vec{z}\big(m^{i}_{B}(x)\big) - \vec{z}\big( {m^{i}_{k}}(x)\big) \rangle
\end{equation}

\com{
\begin{equation} \label{eq:ate}
\widehat{ATE}_d(c_{k}) = \frac{1}{|\mathbf{U_d}|} \sum_{x \in \mathbf{U_d}} f\Big( \vec{z}\big(m^{i}_{B}(x)\big), \vec{z}\big( {m^{i}_{k}}(x)\big)\Big)
\end{equation}
}
where the operator $\langle \rangle$ denotes the total variation distance: A summation over the absolute values of vector coordinates.
\footnote{For a three class prediction and a single example, where the probability distributions for the base and the compressed models are $(0.7, 0.2, 0.1)$ and $(0.5, 0.1, 0.4)$, respectively, \textit{$\widehat{ATE}_i(c_{k})$} = $|0.7-0.5|+|0.2-0.1|+|0.1-0.4|=0.6$.} 
As we are interested in the effect on the probability assigned to each class by the classifier $m^{i}_{k}$, we measure the class probability of its output for an example $x$, as proposed by \citet{feder2021causalm}.\footnote{For sequence tagging tasks, we first compute sentence-level ATEs by averaging the word-level probability differences, and then average those ATEs to get the final ATE.}

In our regression model we choose to include the ATE of the source and the target domains, $\widehat{ATE}_{S^i}(c_{k})$ (estimated on $\mathbf{U_{S^{i}}}$) and $\widehat{ATE}_{T^i}(c_{k})$ (estimated on $\mathbf{U_{T^{i}}}$) , respectively. We note that in computing the ATE we only require the predictions of the models, and do not need  labeled data.

\paragraph{In-domain Performance}
A common metric for selecting a classification model is its performance on a held-out set. Indeed, in cases where we do not have access to any information from the target domain, the naive choice is the best performing model on a held-out source domain set \cite{elsahar2019annotate}. Hence, for every $c_{k} \in \mathcal{C}$ we compute the performance of ${m^{i}_{k}}$ on $\mathbf{H_{S^{i}}}$.

\paragraph{Domain Classification}
An important variable when predicting model performance on an unseen test domain is the distance between its training domain and that test domain \cite{elsahar2019annotate}. While there are many ways to approximate this distance, we choose to do so by training a domain classifier on $\mathbf{U_{S^i}}$ and $\mathbf{U_{T^i}}$, classifying each example according to its domain. We then compute the average probability assigned to the target examples to belong to the source domain, according to the domain classifier:
\begin{equation}
    \widehat{P(S^i|T^i)} = \frac{1}{|\mathbf{H_{T^i}}|} \sum_{x \in \mathbf{H_{T^i}}} P(S^i|x),
\end{equation}
where $P(S^i|x)$ denotes for an unlabeled target example $x$, the probability that it belongs to the source domain $\mathbf{S^i}$, based on the domain classifier.

\paragraph{Compression-size Effects}
We include in our regression binary variables indicating the number of layers that were removed. Naturally, we assume that the larger the number of layers removed, the bigger the gap from the base model should be. 

\subsection{Regression Analysis}
\label{subsec:reg}
In order to decide which $c_{k}$ should be removed from the base model, we follow the process described in Algorithm \ref{alg:causal_compression} for all $c \in \mathcal{C}$ and end up with many candidate compressed models, differing by the model components that were removed. As our goal is to choose a candidate model to be used in an unseen target domain, we train a standard linear stepwise regression model \cite{hocking1976biometrics, draper1998applied, dubossarsky2020secret} to predict the candidate's performance on the seen target domains: 
\begin{equation}
\label{eq:reg}
    Y = \beta_0 + \beta_1 X_1 + \cdots + \beta_m X_m + \epsilon,
\end{equation}
where $Y$ is performance on these target domains, computed using their held-out sets $(\mathbf{H_{T^i}})_{i=1}^{n}$, and ${X_1, \cdots, X_m}$ are the set of variables described in \ref{subsec:features}, including the ATE. In stepwise regression variables are added to the model incrementally only if their marginal addition for predicting $Y$ is statistically significant ($p < 0.01$). This method is useful for finding variables with maximal and unique contribution to the explanation of $Y$. The value of this regression is two-fold in our case as it allows us to: (1) get a predictive model that can choose a high quality compressed model candidate, and (2) estimate the predictive power of the ATE on model performance in the target domain.


\section{Experiments}
\label{sec:experiments}
\subsection{Data}

We consider three challenging datasets (tasks):\\
\textbf{(1)} The Amazon product reviews dataset for sentiment classification ~\cite{he2016ups}.\footnote{\url{http://jmcauley.ucsd.edu/data/amazon/}.} This dataset consists of product reviews and metadata, from which we choose 6 distinct domains: Amazon Instant Video (AIV), Beauty (B), Digital Music (DM), Musical Instruments (MI), Sports and Outdoors (SAO) and Video Games (VG). All reviews are annotated with an integer score between 0 and 5. We label $> 3$ reviews as positive and $< 3$ reviews as negative. Ambiguous reviews (rating $=3$) are discarded. Since the dataset does not contain development and test sets, we randomly split each domain into training (64\%), development (16\%) and test (20\%) sets. 

\textbf{(2)} The Multi-Genre Natural Language Inference (MultiNLI) corpus for natural language inference classification \cite{williams2018broad} .\footnote{\url{https://cims.nyu.edu/~sbowman/multinli/}.} This corpus consists of pairs of sentences, a premise and a hypothesis, where the hypothesis either entails the premise, is neutral to it or contradicts it. The MultiNLI dataset extends upon the SNLI  corpus ~\cite{bowman2015large}, assembled from image captions, to 10 additional domains: 5 $matched$ domains, containing training, development and test samples and 5 $mismatched$, containing only development and test samples. We experiment with the original SNLI corpus (Captions domain) as well as the $matched$ version of MultiNLI, containing the Fiction, Government, Slate, Telephone and Travel domains, for a total of 6 domains.

\textbf{(3)} The OntoNotes 5.0 dataset \cite{hovy2006ontonotes}, consisting of sentences annotated with named entities, part-of-speech tags and parse trees. \footnote{\url{https://catalog.ldc.upenn.edu/LDC2013T19}.} We focus on the Named Entity Recognition (NER) task with 6 different English domains: Broadcast Conversation (BC), Broadcast News (BN), Magazine (MZ), Newswire (NW), Telephone Conversation (TC) and Web data (WB). This setup allows us to evaluate the quality of \textbf{AMoC} on a sequence tagging task.

The statistics of our experimental setups are reported in Table \ref{table:dataset_statistics}. Since the test sets of the MultiNLI domains are not publicly available, we treat the original development sets as our test sets, and randomly choose 2,000 examples from the training set of each domain to serve as the development sets. We use the original splits of the SNLI as they are all publicly available. Since our datasets manifest class imbalance phenomena we use the macro average F1 as our evaluation measure. 

For the regression step of Algorithm \ref{alg:causal_compression}, we use the development set of each target domain to compute the model's macro F1 score (for the $Y$ and the in-domain performance variables). We compute the ATE variables on the development sets of both domains, train the domain classifier on unlabeled versions of the training sets and compute $\widehat{P(S|T)}$ on the target development set.

\begin{table}[!t]
   \scalebox{0.5}{
\begin{tabular}{|p{0.3\textwidth}|p{0.18\textwidth}|p{0.18\textwidth}|p{0.18\textwidth}|}
\hline
\multicolumn{4}{|c|}{\textbf{Amazon Reviews}}                        \\ \hline
                    & \textbf{Train} & \textbf{Dev}  & \textbf{Test} \\ \hline
\textbf{Amazon Instant Video}        & \textbf{21K}   & \textbf{5.2K} & \textbf{6.5K} \\ \hline
\textbf{Beauty}          & \textbf{112K}  & \textbf{28K}  & \textbf{35K}  \\ \hline
\textbf{Digital Music}         & \textbf{37K}   & \textbf{9.2K} & \textbf{11K}  \\ \hline
\textbf{Musical Instruments}         & \textbf{6K}    & \textbf{1.5K} & \textbf{1.9K} \\ \hline
\textbf{Sports and Outdoors}        & \textbf{174K}  & \textbf{43K}  & \textbf{54K}  \\ \hline
\textbf{Video Games}         & \textbf{130K}  & \textbf{32K}  & \textbf{40K}  \\ \hline
\multicolumn{4}{|c|}{\textbf{MultiNLI}}                              \\ \hline
                    & \textbf{Train} & \textbf{Dev}  & \textbf{Test} \\ \hline
\textbf{Captions}   & \textbf{550K}  & \textbf{10K}  & \textbf{10K}  \\ \hline
\textbf{Fiction}    & \textbf{75K}   & \textbf{2K}   & \textbf{2K}   \\ \hline
\textbf{Government} & \textbf{75K}   & \textbf{2K}   & \textbf{2K}   \\ \hline
\textbf{Slate}      & \textbf{75K}   & \textbf{2K}   & \textbf{2K}   \\ \hline
\textbf{Telephone}  & \textbf{81K}   & \textbf{2K}   & \textbf{2K}   \\ \hline
\textbf{Travel}     & \textbf{75K}   & \textbf{2K}   & \textbf{2K}   \\ \hline
\multicolumn{4}{|c|}{\textbf{OntoNotes}}                             \\ \hline
                    & \textbf{Train} & \textbf{Dev}  & \textbf{Test} \\ \hline
\textbf{Broadcast Conversation}         & \textbf{173K}  & \textbf{30K}  & \textbf{36K}  \\ \hline
\textbf{Broadcast News}         & \textbf{207K}  & \textbf{25K}  & \textbf{26K}  \\ \hline
\textbf{Magazine}         & \textbf{161K}  & \textbf{15K}  & \textbf{17K}  \\ \hline
\textbf{News}         & \textbf{878K}  & \textbf{148K} & \textbf{60K}  \\ \hline
\textbf{Telephone Conversation}         & \textbf{92K}   & \textbf{11K}  & \textbf{11K}  \\ \hline
\textbf{Web}         & \textbf{361K}  & \textbf{48K}  & \textbf{50K}  \\ \hline
\end{tabular}}
    \caption{Data statistics. We report the number of sentences for Amazon Reviews and MultiNLI, and the number of tokens for OntoNotes.}
    \label{table:dataset_statistics}
\end{table}

\subsection{Model and Baselines}
\paragraph{Model}
\label{subsec:arch}

The encoder being compressed is the BERT-base model \cite{devlin2019bert}. BERT is a 12-layer Transformer model \citet{vaswani2017attention, radford2018improving}, representing textual inputs contextually and sequentially. Our decoder consists of a layer attention mechanism \cite{kondratyuk201975} which computes a parameterized weighted average over the layers' output, followed by a $1D$ convolution with the max-pooling operation and a final Softmax layer. Figure \ref{fig:model_architecture}(a) presents a simplified version of the architecture of this model with 3 encoder layers.

\paragraph{Baselines}
To put our results in context of previous model compression work, we compare our models to three strong baselines. Like \textbf{AMoC}, the baselines generate reduced-size encoders. These encoders are augmented with the same decoder as in our model to yield the baseline architectures.

The first baseline is \textbf{DistilBERT} (\textbf{DB}) \cite{sanh2019distilbert}: A 6-layer compressed version of BERT-base, trained on the masked language modelling task with the goal of mimicking the predictions of the larger model. We used its default setting, i.e., removal of 6 layers with $c=\{2, 4, 6, 7, 9, 11\}$. \citet{sanh2019distilbert} demonstrated that \textbf{DistilBERT} achieves comparable results to the large model with only half of its layers.

Since \textbf{DistilBERT} was not designed or tested on out-of-distribution data, we create an additional version, denoted as \textbf{DB + DA}. In this version, the training process is performed on the masked language modelling task using an unlabeled version of the training data from both the source and the target domains, with its original hyper-parameters. 

We further add an additional adaptation-aware baseline: \textbf{DB + GR}, the \textbf{DistilBERT} model equipped with the gradient reversal (GR) layer \cite{ganin2015unsupervised}. Particularly, we augment the \textbf{DistilBERT} model with a domain classifier, similar in structure to the task classifier, which aims to distinguish between the unlabeled source and the unlabeled target examples. By reversing the gradients resulting from the objective function of this classifier, the encoder is biased to produce domain-invariant representations. We set the weights of the main task loss and the domain classification loss to $1$ and $0.1$, respectively.

Another baseline is \textbf{LayerDrop} (\textbf{LD}), a procedure that applies layer dropout during training, making the model robust to the removal of certain layers during inference \cite{fan2019reducing}. During training, we apply a fixed dropout rate of $0.5$ for all layers. At inference, we apply their $Every$ $Other$ strategy by removing all even layers to obtain a reduced 6-layer model. 

Finally, we compare \textbf{AMoC} to \textbf{ALBERT}, a recently proposed BERT-based variant designed to mimic the performance of the larger BERT model with only a tenth of its parameters (11M parameters compared to BERT's 110M parameters) \cite{lan2020albert}. \textbf{ALBERT} is trained with cross-layer parameter sharing and sentence ordering objectives, leading to better model efficiency. Unlike other baselines explored here, it is not directly comparable since it consists of 12 layers and was pre-trained on substantially more data. As such, we do not include it in the main results table (Table \ref{table:domain_adaptation}), and instead discuss its performance compared to \textbf{AMoC} in Section \ref{sec:results}.

\subsection{Compression Scheme Experiments}

While our compression algorithm is neither restricted to a specific DNN architecture nor to the removal of certain model components, we follow previous work and focus on the removal of layer sets \cite{fan2019reducing, sanh2019distilbert, sajjad2020poor}. With the goal of addressing our research questions posed in $\S$~\ref{sec:intro}, we perform extensive compression experiments on the 12-layer BERT by considering the removal of $4, 6$ and $8$ layers. For each number of layers removed, we randomly sample $100$ layer sets to generate our model candidates. To be able to test our method on all domain pairs, we randomly split these pairs into five 20\% domain pair sets and train five regression models, differing in the set used for testing. Our splits respect the restriction that no test set domain (source or target) appears in the training set.

\subsection{Hyper-parameters}
We implement all models using HuggingFace's Transformers package \cite{wolf2019huggingface}.\footnote{\url{https://github.com/huggingface/transformers}.} We consider the following hyper-parameters for the uncompressed models: Training for 10 epochs (Amazon Reviews and MultiNLI) or 30 epochs (OntoNotes) with an early stopping criterion according to the development set, optimizing all parameters using the ADAM optimizer \cite{kingma2015adam} with a weight decay of 0.01 and a learning rate of 1e-4, a batch size of 32, a window size of 9, 16 output channels for the $1D$ convolution, and a dropout layer probability of 0.1 for the layer attention module. The compressed models are trained on the labeled source data for 1 epoch (Amazon Reviews and MultiNLI) or 10 epochs (OntoNotes).

The domain classifiers are identical in architecture to our task classifiers and use the uncompressed encoder after it was optimized during the above task-based training. These classifiers are trained on the unlabeled version of the source and target training sets for 25 epochs with early stopping, using the same hyper-parameters as above.

\section{Results}
\label{sec:results}
\begin{figure}
\centering
\includegraphics[width=0.99\linewidth]{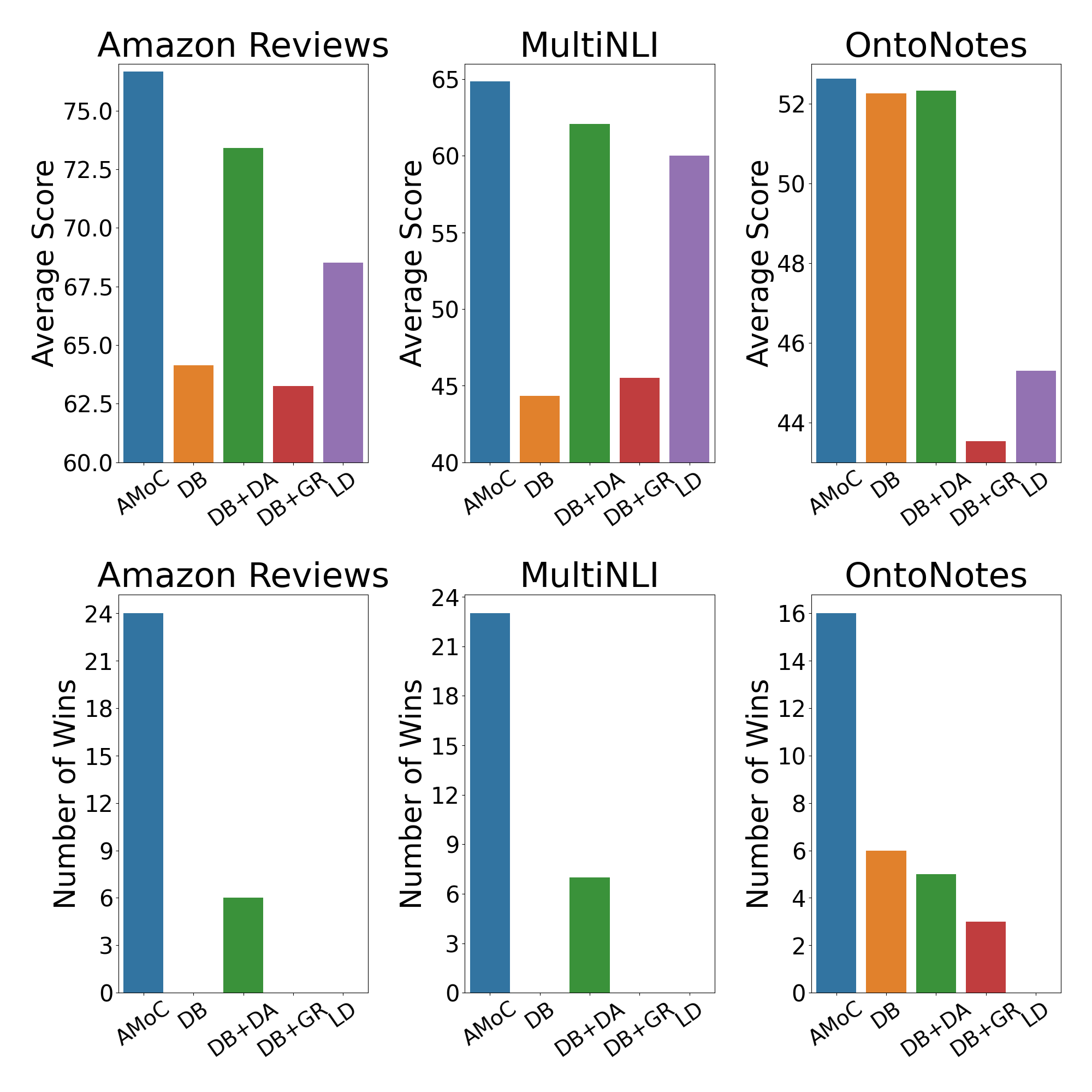}
\caption{Summary of domain adaptation results. Overall average score (up) and overall number of wins (down) over all source-target domain pairs.}
\label{fig:main_results_summary}
\end{figure}

\paragraph{Performance of Compressed Models}
Table \ref{table:domain_adaptation} reports macro F1 scores for all domain pairs of the Amazon Reviews, MultiNLI and OntoNotes datasets, when considering the removal of 6 layers, while Figure \ref{fig:main_results_summary} provides summary statistics. Clearly, \textbf{AMoC} outperforms all baselines in the vast majority of setups (see, e.g., the lower graphs of Figure \ref{fig:main_results_summary}). Moreover, its average target-domain performance (across the 5 source domains) improves over the second best model (\textbf{DB + DA}) by up to 4.56\%, 5.16\% and 1.63\%, on Amazon Reviews, MultiNLI and OntoNotes, respectively (lowest rows of each table in Table \ref{table:domain_adaptation}; see also the average across setups in the upper graphs of Figure \ref{fig:main_results_summary}). These results provide a positive answer to Q$1$ of $\S$ \ref{sec:intro}, by indicating the superiority of \textbf{AMoC} over strong alternatives.

\textbf{DB+GR} is overall the worst performing baseline, followed by \textbf{DB}, with an average degradation of \com{6.3\%} 11.3\% and 8.2\% macro F1 score, respectively, compared to the more successful cross-domain oriented variant \textbf{DB + DA}. This implies that out-of-the-box compressed models such as \textbf{DB} struggle to generalize well to out-of-distribution data. \textbf{DB + DA} also performs worse than \textbf{AMoC} in a large portion of the experiments. These results are even more appealing given that \textbf{AMoC} does not perform any gradient step on the target data, performing only a small number of gradient steps on the source data. In fact, \textbf{AMoC} only uses the unlabeled target data for computing the regression features. Lastly, \textbf{LD}, another strong baseline which was specifically designed to remove layers from BERT, is surpassed by \textbf{AMoC} by as much as \com{8.18\% \com{8\%} and 4.7\% on Amazon Reviews and MultiNLI, respectively} 6.76\% F1, when averaging over all source-target domain pairs.


Finally, we compare \textbf{AMoC} to \textbf{ALBERT}. We find that on average \textbf{ALBERT} is outperformed by \textbf{AMoC} by $8.8 \%$ F1 on Amazon Reviews, and by $1.6 \%$ F1 on MultiNLI. On OntoNotes the performance gap between \textbf{ALBERT} and \textbf{AMoC} is an astounding $24.8 \%$ F1 in favor of \textbf{AMoC}, which might be a result of \textbf{ALBERT} being an uncased model, an important feature for NER tasks.

\begin{table*}[!ht]
\scalebox{0.55}{
\begin{tabular}{|c|c|c|c|c|c|c|c|c|c|c|c|c|c|c|c|c|c|c|}
\hline
\multicolumn{19}{|c|}{\textbf{Amazon Reviews}}                                                                                                                                                                                                                                                                                                                                                \\ \hline
\textbf{S\textbackslash{}T} & \multicolumn{6}{c|}{\textbf{AIV}}                                                                                 & \multicolumn{6}{c|}{\textbf{B}}                                                                                         & \multicolumn{6}{c|}{\textbf{DM}}                                                                                  \\ \hline
                            & \textbf{Base} & \textbf{AMoC}        & \textbf{DB}          & \textbf{DB+DA}       & \textbf{DB+GR} & \textbf{LD} & \textbf{Base} & \textbf{AMoC}        & \textbf{DB}          & \textbf{DB+DA}       & \textbf{DB+GR}       & \textbf{LD} & \textbf{Base} & \textbf{AMoC}        & \textbf{DB}          & \textbf{DB+DA}       & \textbf{DB+GR} & \textbf{LD} \\ \hline
\textbf{AIV}                &               &                      &                      &                      &                &             & 75.49         & {\ul \textbf{82.14}} & 65.00                & {\ul 75.86}          & 65.42                & 69.51       & 77.66         & \textbf{76.02}       & 67.12                & 75.94                & 62.8           & 71.92       \\ \hline
\textbf{B}                  & 80.05         & \textbf{79.18}       & 69.23                & 74.07                & 66.73          & 74.10       &               &                      &                      &                      &                      &             & 77.10         & \textbf{76.60}       & 65.42                & 72.74                & 58.52          & 69.94       \\ \hline
\textbf{DM}                 & 78.97         & \textbf{78.57}       & 69.52                & 76.00                & 70.39          & 72.14       & 76.54         & 74.37                & 63.83                & \textbf{74.94}       & 65.21                & 67.36       &               &                      &                      &                      &                &             \\ \hline
\textbf{MI}                 & 65.24         & {\ul \textbf{69.87}} & 54.96                & {\ul 67.21}          & 55.99          & 56.53       & 72.72         & {\ul 72.78}          & 55.75                & {\ul \textbf{74.83}} & 46.44                & 61.25       & 60.09         & {\ul 63.88}          & 50.01                & {\ul \textbf{68.24}} & 30.42          & 52.67       \\ \hline
\textbf{SAO}                & 77.10         & {\ul \textbf{77.64}} & 63.26                & 70.01                & 63.43          & 67.72       & 83.88         & {\ul \textbf{85.12}} & 69.87                & 81.74                & 67.19                & 76.32       & 74.30         & {\ul \textbf{75.15}} & 58.51                & 67.60                & 60.58          & 64.60       \\ \hline
\textbf{VG}                 & 82.73         & {\ul \textbf{83.79}} & 73.66                & 78.98                & 73.24          & 76.24       & 85.20         & {\ul \textbf{85.21}} & 69.62                & 80.34                & 70.91                & 77.13       & 81.10         & {\ul \textbf{82.43}} & 71.21                & 75.08                & 72.51          & 76.01       \\ \hline
\textbf{AVG}                & 76.81         & {\ul \textbf{77.81}} & 66.13                & 73.25                & 65.96          & 69.35       & 78.77         & {\ul \textbf{79.92}} & 64.81                & 77.54                & 63.03                & 70.31       & 74.05         & {\ul \textbf{74.82}} & 62.45                & 71.92                & 56.97          & 67.03       \\ \hline
                            & \multicolumn{6}{c|}{\textbf{MI}}                                                                                  & \multicolumn{6}{c|}{\textbf{SAO}}                                                                                       & \multicolumn{6}{c|}{\textbf{VG}}                                                                                  \\ \hline
                            & \textbf{Base} & \textbf{AMoC}        & \textbf{DB}          & \textbf{DB+DA}       & \textbf{DB+GR} & \textbf{LD} & \textbf{Base} & \textbf{AMoC}        & \textbf{DB}          & \textbf{DB+DA}       & \textbf{DB+GR}       & \textbf{LD} & \textbf{Base} & \textbf{AMoC}        & \textbf{DB}          & \textbf{DB+DA}       & \textbf{DB+GR} & \textbf{LD} \\ \hline
\textbf{AIV}                & 67.99         & \textbf{64.44}       & 58.26                & 61.64                & 58.64          & 61.43       & 69.76         & 69.52                & 59.71                & {\ul \textbf{71.62}} & 58.96                & 62.97       & 77.71         & \textbf{76.52}       & 67.43                & 76.44                & 67.11          & 70.19       \\ \hline
\textbf{B}                  & 82.70         & \textbf{80.16}       & 66.47                & 76.28                & 68.03          & 71.87       & 83.73         & \textbf{83.21}       & 72.23                & 79.57                & 72.11                & 77.29       & 82.57         & \textbf{82.23}       & 65.52                & 76.96                & 65.50          & 71.59       \\ \hline
\textbf{DM}                 & 71.53         & \textbf{71.10}       & 59.18                & 67.21                & 61.37          & 63.13       & 70.94         & 63.83                & 58.45                & \textbf{65.29}       & 61.75                & 62.79       & 78.45         & 76.04                & 68.67                & \textbf{76.21}       & 66.93          & 70.66       \\ \hline
\textbf{MI}                 &               &                      &                      &                      &                &             & 70.08         & {\ul \textbf{72.71}} & 59.23                & {\ul 71.39}          & 58.30                & 66.10       & 65.10         & {\ul \textbf{67.91}} & 51.60                & 56.37                & 49.67          & 56.87       \\ \hline
\textbf{SAO}                & 84.16         & {\ul \textbf{84.73}} & 71.09                & 78.64                & 72.27          & 72.44       &               &                      &                      &                      &                      &             & 80.05         & {\ul \textbf{81.06}} & 64.51                & 75.14                & 65.78          & 70.00       \\ \hline
\textbf{VG}                 & 86.43         & \textbf{82.07}       & 66.22                & 76.77                & 67.38          & 70.59       & 82.61         & \textbf{82.23}       & 68.96                & 79.12                & 70.18                & 73.83       &               &                      &                      &                      &                &             \\ \hline
\textbf{AVG}                & 78.56         & \textbf{76.50}       & 64.24                & 72.11                & 65.54          & 67.89       & 75.42         & \textbf{74.30}       & 63.72                & 73.40                & 64.26                & 68.60       & 76.78         & \textbf{76.75}       & 63.55                & 72.22                & 63.00          & 67.86       \\ \hline
\multicolumn{19}{|c|}{\textbf{MNLI}}                                                                                                                                                                                                                                                                                                                                                          \\ \hline
\textbf{S\textbackslash{}T} & \multicolumn{6}{c|}{\textbf{Captions}}                                                                            & \multicolumn{6}{c|}{\textbf{Fiction}}                                                                                   & \multicolumn{6}{c|}{\textbf{Govern.}}                                                                             \\ \hline
                            & \textbf{Base} & \textbf{AMoC}        & \textbf{DB}          & \textbf{DB+DA}       & \textbf{DB+GR} & \textbf{LD} & \textbf{Base} & \textbf{AMoC}        & \textbf{DB}          & \textbf{DB+DA}       & \textbf{DB+GR}       & \textbf{LD} & \textbf{Base} & \textbf{AMoC}        & \textbf{DB}          & \textbf{DB+DA}       & \textbf{DB+GR} & \textbf{LD} \\ \hline
\textbf{Captions}           &               &                      &                      &                      &                &             & 58.92         & \textbf{58.37}       & 46.96                & 57.37                & 46.04                & 54.93       & 59.51         & \textbf{59.35}       & 40.14                & 57.85                & 42.54          & 56.85       \\ \hline
\textbf{Fiction}            & 71.33         & \textbf{68.81}       & 39.04                & 67.60                & 45.26          & 63.27       &               &                      &                      &                      &                      &             & 73.41         & \textbf{69.71}       & 46.83                & 69.55                & 47.10          & 63.56       \\ \hline
\textbf{Govern.}            & 62.52         & {\ul \textbf{68.04}} & 44.45                & {\ul 63.47}          & 39.23          & 54.68       & 67.61         & \textbf{66.05}       & 44.5                 & 63.47                & 46.75                & 60.44       &               &                      &                      &                      &                &             \\ \hline
\textbf{Slate}              & 65.04         & \textbf{62.40}       & 37.58                & 46.99                & 44.87          & 55.39       & 69.83         & \textbf{67.70}       & 46.53                & 58.59                & 43.58                & 62.07       & 72.95         & \textbf{72.16}       & 49.53                & 71.31                & 49.23          & 66.82       \\ \hline
\textbf{Telephone}          & 65.04         & \textbf{61.22}       & 40.03                & 58.65                & 36.64          & 59.77       & 69.07         & \textbf{67.77}       & 47.45                & 63.76                & 46.76                & 61.91       & 71.46         & 65.47                & 46.83                & \textbf{66.63}       & 45.99          & 65.53       \\ \hline
\textbf{Travel}             & 65.77         & \textbf{62.11}       & 36.54                & 60.11                & 38.29          & 55.41       & 66.97         & \textbf{65.19}       & 44.05                & 60.06                & 42.94                & 56.67       & 74.24         & 72.07                & 49.03                & \textbf{72.69}       & 51.32          & 65.47       \\ \hline
\textbf{AVG}                & 65.94         & \textbf{64.52}       & 39.53                & 59.36                & 40.86          & 57.70       & 66.48         & \textbf{65.02}       & 45.90                & 60.65                & 45.21                & 59.20       & 70.31         & \textbf{67.75}       & 46.47                & 67.61                & 47.24          & 63.65       \\ \hline
                            & \multicolumn{6}{c|}{\textbf{Slate}}                                                                               & \multicolumn{6}{c|}{\textbf{Telephone}}                                                                                 & \multicolumn{6}{c|}{\textbf{Travel}}                                                                              \\ \hline
                            & \textbf{Base} & \textbf{AMoC}        & \textbf{DB}          & \textbf{DB+DA}       & \textbf{DB+GR} & \textbf{LD} & \textbf{Base} & \textbf{AMoC}        & \textbf{DB}          & \textbf{DB+DA}       & \textbf{DB+GR}       & \textbf{LD} & \textbf{Base} & \textbf{AMoC}        & \textbf{DB}          & \textbf{DB+DA}       & \textbf{DB+GR} & \textbf{LD} \\ \hline
\textbf{Captions}           & 52.83         & {\ul \textbf{53.26}} & 41.30                & 52.96                & 42.23          & 50.56       & 56.94         & 56.68                & 41.22                & {\ul \textbf{58.35}} & 45.53                & 54.01       & 57.88         & \textbf{57.40}       & 42.86                & 54.84                & 43.64          & 54.88       \\ \hline
\textbf{Fiction}            & 66.76         & 62.94                & 44.79                & \textbf{64.13}       & 45.70          & 59.82       & 71.83         & \textbf{68.47}       & 41.66                & 67.70                & 44.52                & 64.97       & 69.86         & 66.28                & 46.98                & \textbf{66.52}       & 46.81          & 62.36       \\ \hline
\textbf{Govern.}            & 65.22         & {\ul \textbf{65.59}} & 46.57                & 62.89                & 45.42          & 61.06       & 67.54         & {\ul \textbf{67.87}} & 43.73                & 65.39                & 45.88                & 65.46       & 67.45         & 64.70                & 48.67                & \textbf{66.99}       & 48.58          & 63.09       \\ \hline
\textbf{Slate}              &               &                      &                      &                      &                &             & 68.27         & {\ul \textbf{71.27}} & 45.21                & 59.39                & 39.50                & 61.06       & 71.47         & \textbf{69.01}       & 46.19                & 57.94                & 46.92          & 61.79       \\ \hline
\textbf{Telephone}          & 65.53         & \textbf{63.62}       & 45.73                & 56.35                & 44.68          & 60.70       &               &                      &                      &                      &                      &             & 69.20         & \textbf{65.97}       & 47.30                & 65.53                & 42.94          & 61.76       \\ \hline
\textbf{Travel}             & 65.02         & 60.11                & 45.65                & \textbf{60.96}       & 47.08          & 56.51       & 69.57         & \textbf{66.31}       & 42.30                & 64.35                & 45.86                & 61.63       &               &                      &                      &                      &                &             \\ \hline
\textbf{AVG}                & 63.07         & \textbf{61.10}       & 44.81                & 59.46                & 45.02          & 57.73       & 66.83         & \textbf{66.12}       & 42.82                & 63.04                & 44.26                & 61.43       & 67.17         & \textbf{64.67}       & 46.40                & 62.36                & 45.78          & 60.78       \\ \hline
\multicolumn{19}{|c|}{\textbf{OntoNotes}}                                                                                                                                                                                                                                                                                                                                                     \\ \hline
\textbf{S\textbackslash{}T} & \multicolumn{6}{c|}{\textbf{BC}}                                                                                  & \multicolumn{6}{c|}{\textbf{BN}}                                                                                        & \multicolumn{6}{c|}{\textbf{MZ}}                                                                                  \\ \hline
\textbf{}                   & \textbf{Base} & \textbf{AMoC}        & \textbf{DB}          & \textbf{DB+DA}       & \textbf{DB+GR} & \textbf{LD} & \textbf{Base} & \textbf{AMoC}        & \textbf{DB}          & \textbf{DB+DA}       & \textbf{DB+GR}       & \textbf{LD} & \textbf{Base} & \textbf{AMoC}        & \textbf{DB}          & \textbf{DB+DA}       & \textbf{DB+GR} & \textbf{LD} \\ \hline
\textbf{BC}                 &               &                      &                      &                      &                &             & 73.78         & \textbf{71.28}       & 70.76                & 70.94                & 58.22                & 66.46       & 64.06         & 60.96                & 63.44                & {\ul \textbf{64.75}} & 48.48          & 53.78       \\ \hline
\textbf{BN}                 & 74.25         & \textbf{71.06}       & 70.83                & 70.11                & 70.29          & 65.61       &               &                      &                      &                      &                      &             & 69.92         & 68.34                & 68.71                & 69.39                & \textbf{69.70} & 60.87       \\ \hline
\textbf{MZ}                 & 66.56         & 62.00                & 60.55                & 61.76                & \textbf{62.06} & 54.76       & 71.47         & \textbf{67.32}       & 66.5                 & 66.41                & 59.67                & 61.29       &               &                      &                      &                      &                &             \\ \hline
\textbf{NW}                 & 72.23         & \textbf{70.26}       & 68.22                & 70.16                & 41.20          & 63.57       & 80.85         & \textbf{79.54}       & 78.15                & 79.34                & 68.92                & 75.07       & 74.66         & 71.78                & 71.86                & \textbf{72.28}       & 65.76          & 64.82       \\ \hline
\textbf{TC}                 & 42.63         & 41.78                & {\ul \textbf{45.14}} & 39.18                & 21.32          & 29.64       & 53.08         & 52.37                & {\ul \textbf{54.56}} & 51.69                & 19.80                & 42.16       & 39.17         & 38.59                & {\ul \textbf{41.94}} & 38.75                & 16.98          & 33.81       \\ \hline
\textbf{WB}                 & 28.47         & \textbf{27.58}       & 26.79                & 25.17                & 26.97          & 21.97       & 40.39         & {\ul \textbf{40.68}} & 39.09                & 40.35                & 30.79                & 33.55       & 15.86         & {\ul 20.09}          & {\ul 22.84}          & {\ul \textbf{24.84}} & 15.53          & 13.52       \\ \hline
\textbf{AVG}                & 56.83         & \textbf{54.54}       & 54.31                & 53.28                & 44.37          & 47.11       & 63.91         & \textbf{62.24}       & 61.81                & 61.75                & 47.48                & 55.71       & 52.73         & 51.95                & {\ul 53.76}          & {\ul \textbf{54.00}} & 43.29          & 45.36       \\ \hline
                            & \multicolumn{6}{c|}{\textbf{NW}}                                                                                  & \multicolumn{6}{c|}{\textbf{TC}}                                                                                        & \multicolumn{6}{c|}{\textbf{WB}}                                                                                  \\ \hline
                            & \textbf{Base} & \textbf{AMoC}        & \textbf{DB}          & \textbf{DB+DA}       & \textbf{DB+GR} & \textbf{LD} & \textbf{Base} & \textbf{AMoC}        & \textbf{DB}          & \textbf{DB+DA}       & \textbf{DB+GR}       & \textbf{LD} & \textbf{Base} & \textbf{AMoC}        & \textbf{DB}          & \textbf{DB+DA}       & \textbf{DB+GR} & \textbf{LD} \\ \hline
\textbf{BC}                 & 61.31         & \textbf{58.80}       & 58.44                & 57.75                & 46.95          & 50.73       & 62.39         & {\ul \textbf{63.07}} & 58.19                & 59.53                & 59.31                & 55.21       & 48.90         & \textbf{47.42}       & 45.58                & 46.00                & 45.56          & 40.17       \\ \hline
\textbf{BN}                 & 73.55         & 69.79                & 70.51                & \textbf{71.26}       & 58.80          & 62.31       & 69.64         & \textbf{65.69}       & 61.45                & 64.68                & 64.98                & 60.40       & 51.34         & \textbf{50.14}       & 48.02                & 48.72                & 48.39          & 43.45       \\ \hline
\textbf{MZ}                 & 67.40         & \textbf{63.80}       & 63.04                & 63.64                & 50.33          & 52.08       & 60.31         & 56.94                & 54.61                & 55.51                & {\ul \textbf{63.37}} & 42.00       & 48.25         & \textbf{44.78}       & 43.11                & 43.91                & 39.98          & 38.80       \\ \hline
\textbf{NW}                 &               &                      &                      &                      &                &             & 61.20         & \textbf{51.88}       & 50.73                & 49.78                & 36.48                & 44.38       & 52.23         & \textbf{50.52}       & 49.07                & 49.30                & 41.34          & 45.72       \\ \hline
\textbf{TC}                 & 35.25         & 35.15                & {\ul \textbf{36.73}} & {\ul 35.58}          & 20.83          & 27.93       &               &                      &                      &                      &                      &             & 36.50         & 35.36                & {\ul \textbf{37.00}} & 36.23                & 25.72          & 27.04       \\ \hline
\textbf{WB}                 & 22.60         & {\ul 26.40}          & {\ul 23.64}          & {\ul \textbf{27.57}} & 20.61          & 17.02       & 18.68         & 15.45                & \textbf{18.36}       & 15.38                & 7.64                 & 10.77       &               &                      &                      &                      &                &             \\ \hline
\textbf{AVG}                & 52.02         & 50.79                & 50.47                & \textbf{51.16}       & 39.50          & 42.01       & 54.44         & \textbf{50.61}       & 48.67                & 48.98                & 46.36                & 42.55       & 47.44         & \textbf{45.64}       & 44.56                & 44.83                & 40.20          & 39.04       \\ \hline
\end{tabular}}
\caption{Domain adaptation results in terms of macro F1 scores on Amazon Reviews (top), MultiNLI (middle) and OntoNotes (bottom) with 6 removed layers. S and T denote Source and Target, respectively. The best result among the compressed models (all models except from Base) is highlighted in bold. We mark results that outperform the uncompressed Base model with an underscore.}
\label{table:domain_adaptation}
\end{table*}

\paragraph{Compressed Model Selection}
We next evaluate how well the regression model and its variables predict the performance of a candidate compressed model on the target domain. Table \ref{tab:r_2} presents the Adjusted $R^2$, indicating the share of the variance in the predicted outcome that the variables explain. Across all experiments and regardless of the number of layers removed, our regression model predicts well the performance on unseen domain pairs, averaging an $R^2$ of $0.881, 0.916$ and $0.826$ on Amazon Reviews, MultiNLI and OntoNotes, respectively. This indicates that our regression properly estimates the performance of candidate models.

\begin{table}[!ht]
    \centering
    \scalebox{0.85}{
    \begin{tabular}{|l|c|c|c|c|} \hline
     & \multicolumn{3}{c|}{$\#$ of removed Layers} &  \\ \hline   
    Dataset & 4  & 6  & 8 & Average \\ \hline   
    Amazon Reviews & 0.844 & 0.898 & 0.902 & 0.881 \\ \hline   
    MultiNLI & 0.902 & 0.921 & 0.926 & 0.916 \\ \hline
    OntoNotes & 0.827 & 0.830 & 0.821 & 0.826 \\ \hline
    \end{tabular}}
    \caption{Adjusted $R^2$ on the test set for each type of compression ($4,6$ or $8$ layers) on each dataset.}
    \label{tab:r_2}
\end{table}

Another support for this observation is that in $75\%$ of the experiments the model selected by the regression is among the top 10 performing compressed candidates. In $55\%$ of the experiments, it is among the top 5 models. On average it performs only $1\%$ worse than the best performing compressed model. Combined with the high adjusted $R^2$ across experiments, this suggests a positive answer to Q$2$ of $\S$ \ref{sec:intro}. 



Finally, as expected, we find that \textbf{AMoC} is often outperformed by the full model. However, the gap between the models is small, averaging only in $1.26\%$ \com{$2.4\%$ in absolute terms}. Moreover, in almost 25\% of all experiments \textbf{AMoC} was able to surpass the full model (underscored scores in Table \ref{table:domain_adaptation}).

\paragraph{Marginal Effects of Regression Variables}
While the performance of the model on data drawn from the same distribution may also be indicative of its out-of-distribution performance, additional information is likely to be needed in order to make an exact prediction. Here, we supplement this indicator with the variables described in $\S$ \ref{subsec:features} and ask whether they can be useful to select the best compressed model out of a set of candidates. Table \ref{tab:stepwise_reg} presents the most statistically significant variables in our stepwise regression analysis. It demonstrates that the ATE and the model's performance in the source domain are usually very indicative of the model's performance.

Indeed, most of the regression's predictive power comes from the model performance on the source domain ($F1_{S}$) and the treatment effects on the source and target domains ($\widehat{ATE_{S}}$, $\widehat{ATE_{T}}$). In contrast, the distance metric ($\widehat{P(S|T})$) and the interaction terms ($\widehat{ATE_{T}}\cdot\widehat{P(S|T)}$, $F1_{S}\cdot\widehat{P(S|T)}$) contribute much less to the total $R^2$. The predictive power of the ATE in both source and target domains suggests a positive answer to Q$3$ of $\S$ \ref{sec:intro}.

\section{Additional Analysis}
\label{sec:analysis}
\subsection{Layer Importance}
\begin{figure*}[!ht]
\centering
\includegraphics[width=0.9\linewidth]{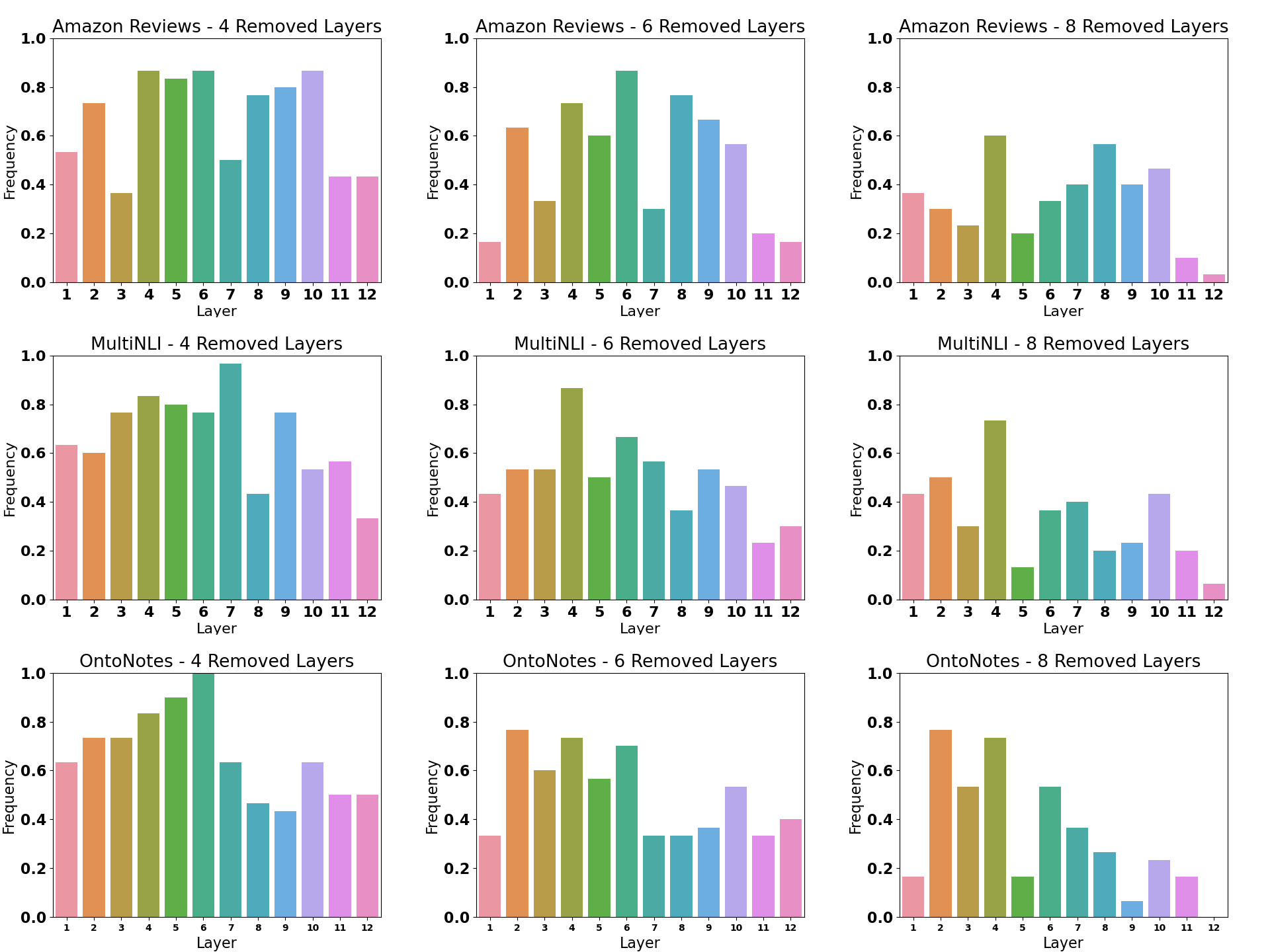}
\caption{Layer frequency at the best (oracle) compressed models when considering the removal of 4, 6 and 8 layers in the three datasets.}
\label{fig:layer_frequency_best}
\end{figure*}

To further understand the importance of each of BERT's layers, we compute the frequency in which each layer appears in the best candidate model, i.e., the model with the highest F1 score on the target test set, of every experiment. Figure \ref{fig:layer_frequency_best} captures the layer frequencies across the different datasets and across the number of removed layers. 

\begin{table}[!t]
    \centering
    \scalebox{0.75}{
    \begin{tabular}{|l|c|c|c|c|c|c|} \hline
     & \multicolumn{2}{|c|}{\textbf{Amazon}} & \multicolumn{2}{|c|}{\textbf{MultiNLI}} & \multicolumn{2}{|c|}{\textbf{OntoNotes}} \\ \hline
    Variable & $\beta$ & $\Delta R^2$ & $\beta$ & $\Delta R^2$ & $\beta$ & $\Delta R^2$ \\ \hline
    $F1_{S}$ & 0.435 & 0.603 & -0.299 & 0.143 & 0.748 & 0.510 \\ \hline
    $\widehat{ATE_{T}}$ & -1.207 & 0.239 & -0.666 & 0.413 & 117.5 & 0.202 \\ \hline
    $\widehat{ATE_{S}}$ & 1.836 & 0.029 & 0.557 & 0.232 & 125.9 & 0.072 \\ \hline
    $\widehat{P(S|T)}$ & -0.298 & 0.028 & -0.652 & 0.061 & 15.60 & 0.052 \\ \hline
    $\widehat{ATE_{T}}$& \multirow{2}{*}{-0.560} & \multirow{2}{*}{0.007} & \multirow{2}{*}{-0.092} & \multirow{2}{*}{0.029} & \multirow{2}{*}{-115.8} & \multirow{2}{*}{0.004} \\
    $\cdot\widehat{P(S|T)}$ & & & & & &  \\ \hline
    $F1_{S}$ & \multirow{2}{*}{0.472} & \multirow{2}{*}{0.004} & \multirow{2}{*}{1.027} & \multirow{2}{*}{0.043} & \multirow{2}{*}{0.187} & \multirow{2}{*}{0.004} \\
    $\cdot\widehat{P(S|T)}$ & & & & & &  \\ \hline
    8 layers & -0.137 & 0.001 & -0.303 & 0.001 & -3.145 & 0.001 \\ \hline
    6 layers & -0.066 & 0 & -0.146 & 0.007 & -1.020 & 0.005 \\ \hline
    const & 0.259 & 0 & 0.594 & 0 & -12.18 & 0 \\ \hline
    \end{tabular}}
    \caption{Stepwise regression coefficients ($\beta$) and their marginal contribution to the adjusted $R^2$ ($\Delta R^2$) on all experiments on both datasets.}
    \label{tab:stepwise_reg}
\end{table}

The plots suggest that the two final layers, layers 11 and 12, are the least important layers with average frequencies of \com{28.8\% and 22.2\%} 30.3\% and 24.8\%, respectively. Additionally, in most cases layer 1 is ranked below the other layers. These results imply that the compressed models are able to better recover from the loss of parameters when the external layers are removed. The most important layer appears to be layer 4, with \com{frequencies of 60.0\% to 83.3\%} an average frequency of 73.3\%. Finally, we notice that a large frequency variance exists across the different subplots. Such variance supports our hypothesis that the decision of which layers to remove should not be based solely on the architecture of the model.


To pin down the importance of a specific layer for a given base model, we utilize a similar regression analysis to that of $\S$ \ref{sec:results}. Specifically, we train a regression model on all compressed candidates for a given source-target domain pair (in all three tasks), adding indicator variables for the exclusion of each layer from the model. This model associates each layer with a regression coefficient, which can be interpreted as the marginal effect of that layer being removed on expected target performance. We then compute for each layer its average coefficient across source-target pairs (Table \ref{tab:layer_importance_reg}, $\beta$ column) and compare it to the fraction of source-target pairs where this layer is not included in the best possible (oracle) compressed model (Table \ref{tab:layer_importance_reg}, $P(\text{Layer removed})$ column). 

As can be seen in the table, layers that their removal is associated with better model performance are more often not included in the best performing compressed models. Indeed, the Spearman's rank correlation between the two rankings is as high as 0.924.  Such analysis demonstrates that the regression model used as part of \textbf{AMoC} not only selects high quality candidates, but can also shed light on the importance of individual layers.


\begin{table}[!ht]
    \centering
    \scalebox{0.75}{
    \begin{tabular}{|c|c|c|} \hline
        Layer Rank & $\bar{\beta}$ & $P(\text{Layer removed})$ \\ \hline
        1 & 0.0448 & 0.300 \\ \hline
        2 & 0.0464 & 0.333 \\ \hline
        3 & 0.0473 & 0.333 \\ \hline
        4 & 0.0483 & 0.333 \\ \hline
        5 & 0.0487 & 0.416 \\ \hline
        6 & 0.0495 & 0.555 \\ \hline
        7 & 0.0501 & 0.472 \\ \hline
        8 & 0.0507 & 0.638 \\ \hline
        9 & 0.0514 & 0.500 \\ \hline
        10 & 0.0522 & 0.638 \\ \hline
        11 & 0.0538 & 0.611 \\ \hline
        12 & 0.0577 & 0.666 \\ \hline
    \end{tabular}}
    \caption{Layer rank according to regression coefficients ($\beta$) and the probability the layer was removed form the best compressed model. Results are averaged across all target-domain pairs in our experiments.}
    \label{tab:layer_importance_reg}
\end{table}

\subsection{Training Epochs}

We next analyze the number of epochs required to fine-tune our compressed models. For each dataset (task) we randomly choose for every target domain 10 compressed models and create two alternatives, differing in the number of training epochs performed after layer removal: One trained for a single epoch and another for 5 epochs (Amazon Reviews, MultiNLI) or 10 epochs (Ontonotes). Table \ref{table:more_epochs_analysis} compares the average F1 (target-domain task performance) and $\widehat{ATE_{T}}$ differences between the two alternatives, on the target domain test and dev sets, respectively. The results suggest that when training for more epochs on Amazon Reviews and MultiNLI the difference in both the F1 and ATE are negligible. For OntoNotes (NER), in contrast, additional training improves the F1, suggesting that further training of the compressed model candidates may be favorable for sequence tagging tasks such as NER.

\begin{table}[!ht]
\centering
\scalebox{0.75}{
\begin{tabular}{|c|c|c|}
\hline
          & F1 Difference & $\widehat{ATE_{T}}$ Difference \\ \hline
Amazon Reviews   & 0.080         & 0.011         \\ \hline
MNLI      & -0.250       & 0.003        \\ \hline
OntoNotes & 2.940         & -0.009        \\ \hline
\end{tabular}}
\caption{F1 and ATE differences when training \textbf{AMoC} after layer removal for multiple epochs vs. a single epoch.}
\label{table:more_epochs_analysis}
\end{table}

\subsection{Space and Time Complexity}

Table \ref{table:num_parameters} compares the number of overall and trainable parameters and the training time of BERT, \textbf{DistilBERT} and \textbf{AMoC}. Removing $L$ layers from BERT yields a reduction of $7L$ million parameters. As can be seen in the Table, \textbf{AMoC} requires training only a small fraction of the overall parameters. Since we only unfreeze one layer per each new connected component, at the worst case our algorithm requires the training of $\min \{L, 12-L\}$ layers. The only exception is in the case where Layer 1 is removed ($1\in c$). In such a case we unfreeze the embedding layer, which adds 24 million trained parameters. In terms of total training time (one epoch of task-based fine-tuning), when averaging over all setups, a single compressed \textbf{AMoC} model is $\times 11$ faster than BERT and $\times 6$ faster than \textbf{DistilBERT}.

\begin{table}[!ht]
\scalebox{0.54}{
\begin{tabular}{|c|c|c|c|}
\hline
\textbf{}     & \textbf{Overall Parameters} & \textbf{Trainable Parameters}                                        & \textbf{Train Time Reduction} \\ \hline
\textbf{BERT-base}  & $110$M & $110$M & $\times 1$    \\ \hline
\textbf{DistilBERT} & $66$M  & $66$M  & $\times 1.83$ \\ \hline
\multirow{2}{*}{\textbf{AMoC}} & \multirow{2}{*}{$110$M - $7$M $\cdot$ L} & $7$M $\cdot$ $\min \{L, 12-L\}$ & \multirow{2}{*}{$\times 11$}                    \\
 & & $+ 17M \cdot \mathbb{1}_{\{1 \in c\}}$ & \\ \hline
\end{tabular}}
\caption{Comparison of number of parameters and training time between BERT-base, \textbf{DistilBERT} and \textbf{AMoC} when removing $L$ layers. \textbf{AMoC}'s number of trainable parameters is an upper bound.}
\label{table:num_parameters}
\end{table}

\subsection{Design Choices}

\paragraph{Computing the ATE} 

Following \citet{goyal2019explaining} and \citet{feder2021causalm}, we implement the ATE with the total variation distance between the probability output of the original model and that of the compressed models. To verify the quality of this design choice, we re-ran our experiments where the ATE is calculated using the KL-divergence between the same distributions. While the results in both conditions are qualitatively similar, we did find a consistent quantitative improvement of the $R^2$  (average of $0.05$ across setups) when considering our total variation distance.

\paragraph{Regression Analysis}

Our regression approach is designed to allow us to both select high-quality compressed candidates and to interpret the importance of each explanatory variable, including the ATEs. As this regression has relatively few features, we do not expect to lose significant predictive power by choosing to focus on linear predictors. To verify this, we re-ran our experiments when using a fully connected feed-forward network \footnote{With one intermediate layer, same input feature as the regression, and hyper-parameters tuned on the development set of each source-target pair.} to predict target performance. This model, which is less interpretable than our regression, is also less accurate: We have observed an increased mean squared error of 1-3\% with the network.

\section{Conclusion}

We explored the relationship between model compression and out-of-distribution generalization. \textbf{AMoC}, our proposed algorithm, relies on causal inference tools for estimating the effects of interventions. It hence creates an interpretable process that allows to understand the role of specific model components. Our results indicate that \textbf{AMoC} is able to produce a smaller model with minimal loss in performance across domains, without any use of target labeled data at test time (Q$1$).

\textbf{AMoC} can efficiently train a large number of compressed model candidates, that can then serve as training examples for a regression model. We have shown that this approach results in a high quality estimation of the performance of compressed models on unseen target domains (Q$2$). Moreover, our stepwise regression analysis indicates that the $\widehat{ATE_{S}}$ and $\widehat{ATE_{T}}$ estimates are instrumental for these attractive properties (Q$3$). 

As training and test set mismatches are common, we steered our model compression research towards out-of-domain generalization. Besides its realistic nature, this setup poses additional modeling challenges, such as understanding the proximity between domains, identifying which components are invariant to domain shift, and estimating performance on unseen domains. Hence, \textbf{AMoC} is designed for model compression in the out-of-distribution setup. We leave the design of similar in-domain compression methods for future work.

Finally, we believe that using causal methods to produce compressed NLP models that can well generalize across distributions is a promising direction of research, and hope that more work will be done in this intersection.

\bibliography{main}
\bibliographystyle{acl_natbib}
\end{document}
